\pdfoutput=1
\documentclass[10pt,twocolumn,letterpaper]{article}

\usepackage{iccv}
\usepackage{times}
\usepackage{epsfig}
\usepackage{graphicx}
\usepackage{amsmath}
\usepackage{amssymb}

\iccvfinalcopy 

\def\iccvPaperID{23} 

\newcommand{\comment}[1]{}

\ificcvfinal\fi
\begin{document}

\title{Triplet-Aware Scene Graph Embeddings}

\author{Brigit Schroeder\thanks{Work done as a resident in Intel AI Lab.} \\
University of California, Santa Cruz\\ 
{\tt\small brschroe@ucsc.edu}
\and
Subarna Tripathi\\
Intel AI Lab\\
{\tt\small subarna.tripathi@intel.com}
\and
Hanlin Tang\\
Intel AI Lab\\
{\tt\small hanlin.tang@intel.com}
}

\maketitle

\label{abs}
\begin{abstract}

 Scene graphs have become an important form of structured knowledge for tasks such as 
for image generation, 
visual relation detection, 
visual question answering, 
 and image retrieval. 
 While visualizing and interpreting word embeddings is well understood, 
 scene graph embeddings have not been fully explored. In this work, we train scene graph embeddings in a layout generation task with different forms of supervision, specifically introducing triplet supervision and data augmentation.  We see a significant performance increase in both metrics that measure the goodness of layout prediction, mean intersection-over-union (mIoU) (52.3\% vs. 49.2\%) and relation score (61.7\% vs. 54.1\%), after the addition of triplet supervision and data augmentation. To understand how these different methods affect the scene graph representation, we apply several new visualization and evaluation methods to explore the evolution of the scene graph embedding. We find that triplet supervision significantly improves the embedding separability, which is highly correlated with the performance of the layout prediction model. 
 
\end{abstract}

\section{Introduction}
\label{sec:intro}
 Scene graphs are a structured data format which encodes semantic relationships between objects \cite{xu2017scenegraph}.  Objects are represented as nodes in the graph and are connected by edges that expresses relationship, in the form of triplets. Each triplet is comprised of a \textit{$<$subject, predicate, object$>$} which semantically describes the relationship between two objects such as \textit{$<$car, on, road$>$} or \textit{$<$dog, left of, person$>$}.
 
Scene graphs can be processed by graph convolutional neural networks (GCNN) which are able to pass information along graph edges\cite{img_gen_from_scene_graph18}. Layout prediction models, such as those used image generation \cite{img_gen_from_scene_graph18}, are multi-stage networks (see Figure \ref{fig:network}) which predict  scene layout masks and object localization (bounding boxes), based upon the structure of the scene graph. The first stage of the layout prediction model is a GCNN that is used to learn a scene graph embedding. Object embeddings are passed to the next stage of the network to predict a bounding box and mask for each object node in the scene graph. We are interested in layout prediction models in particular because a scene graph embedding is learned in the process of training a model. The performance metrics used for evaluating the quality of scene layout, namely mean intersection-over-union (mIoU) and relation score 
\cite{ICLR_LLD},
can be used to correlate with the relative ``quality" of the scene graph embedding. Relation score is particularly relevant as it measures the compliance of the generated layout to the scene graph, possibly giving  insight into how well the embedding is structured, especially in terms of class separability.

In our work, we introduce several new supervisory signals that are conditioned upon triplet embeddings to train scene layout prediction models. We also apply data augmentation by using heuristic-based relationships \cite{ICLR_LLD} to maximize the number of triplets during training. The goal is to learn a triplet-aware scene graph embedding with the hypothesis additional supervision and data augmentation will enrichen the embedding representation.  To this end, we also introduce several methods of embedding visualization and evaluation that help understand the degree of separability that is achieved from our improved layout prediction models. \\

%

\section{Related Work}
\label{sec:related}

Visual Genome \cite{visual_genome16}, a human-annotated scene graph dataset, has known issues of incomplete and incorrect annotations. 
Additionally this only  
provides the bounding boxes for object instances but not their segmentation masks. 
On the other hand, the synthetic scene graphs generated from COCO stuff \cite{cocostuff_caesar2018cvpr} are limited to simple geometric relationships (above, below, left, right, inside, surrounding) but are not hampered by incorrect annotations.  
COCO stuff also provides segmentation masks for instances. 
In this paper we use synthetic scene graphs from COCO stuff \cite{cocostuff_caesar2018cvpr} for our 
experiments. 

Generating scene graphs from visual features \cite{xu2017scenegraph, neural_motifs_2017,graph_RCNN_Yang_2018_ECCV,gjyin_eccv2018, visual_rel_as_func_19, VCTree_Tang_2019_CVPR, graphical_contrastive_loss19, Woo2018LinkNetRE, Qi_2019_CVPR, hr18perminvimg2sg} 
is a relatively explored task. 
Wan \etal \cite{Wan_IJACAI_18} specifically predicts new triplets for scene graph completion using existing scene graphs and visual features. 
Another line of work that emerged recently \cite{img_gen_from_scene_graph18, ICLR_LLD, interactiveGenICLR19, Tripathi2019CompactSG, Jyothi2019LayoutVAESS, ICCV_interactiveGen19, vo2019visualrelation}
take scene graphs as input and produce final RGB images.
All of these methods perform an intermediate layout prediction by learning embeddings of nodes. In closely related work, Belilovsky \etal \cite{belilovsky:hal-01667777} learn a joint visual-scene graph embedding for use in image retrieval.
None of the above explored utilizing \textit{$<$subject, predicate, object$>$} triplets as additional supervisory signal for more effective structured prediction. 

Few papers that aim to detect visual relations from images also analyze the learned embeddings qualitatively. 
Dornadula \etal \cite{visual_rel_as_func_19} visualizes learned object category embeddings 
using 2-dimensional t-SNE plots. 
In Graph-RCNN \cite{graph_RCNN_Yang_2018_ECCV} Yang \etal showed how common sense of \textit{$<$object-predicate$>$} co-occurrence \cite{tutorial_graphrcnn} emerges 
as the top scorers based upon the extracted weights that use visual features. They show similar common sense also appears for \textit{$<$object-object$>$} co-occurrences.  
In the VTransE paper \cite{VTransEZhangKCC17} 
the authors 
utilize t-SNE for visualizing the embeddings of predicate vectors and speculate when the model learns only co-occurrence vs. the actual meaning of the relationship by inspecting the neighbors in the t-SNE space.  
Note all these described models utilize visual features. In contrast, our model is trained from scene graphs without access to 
visual features. We provide quantitative and qualitative measures on \emph{why} the embeddings trained by the proposed model is superior.

From an embedding introspection perspective, the closest work is \cite{obj-glove19}. In this concurrent work, Xu \etal introduced generic embeddings for common visual objects. They visualized embeddings of most common objects using t-SNE, and qualitatively studied results on vector decomposition and projection on specific axes anecdotally. We note that these embeddings are learned solely based on co-occurrence and thus they are not able to deal with relationships between objects.

\section{Method}
\label{sec:method}

\begin{figure*}[ht]
\begin{center}
\includegraphics[width=0.8\textwidth]{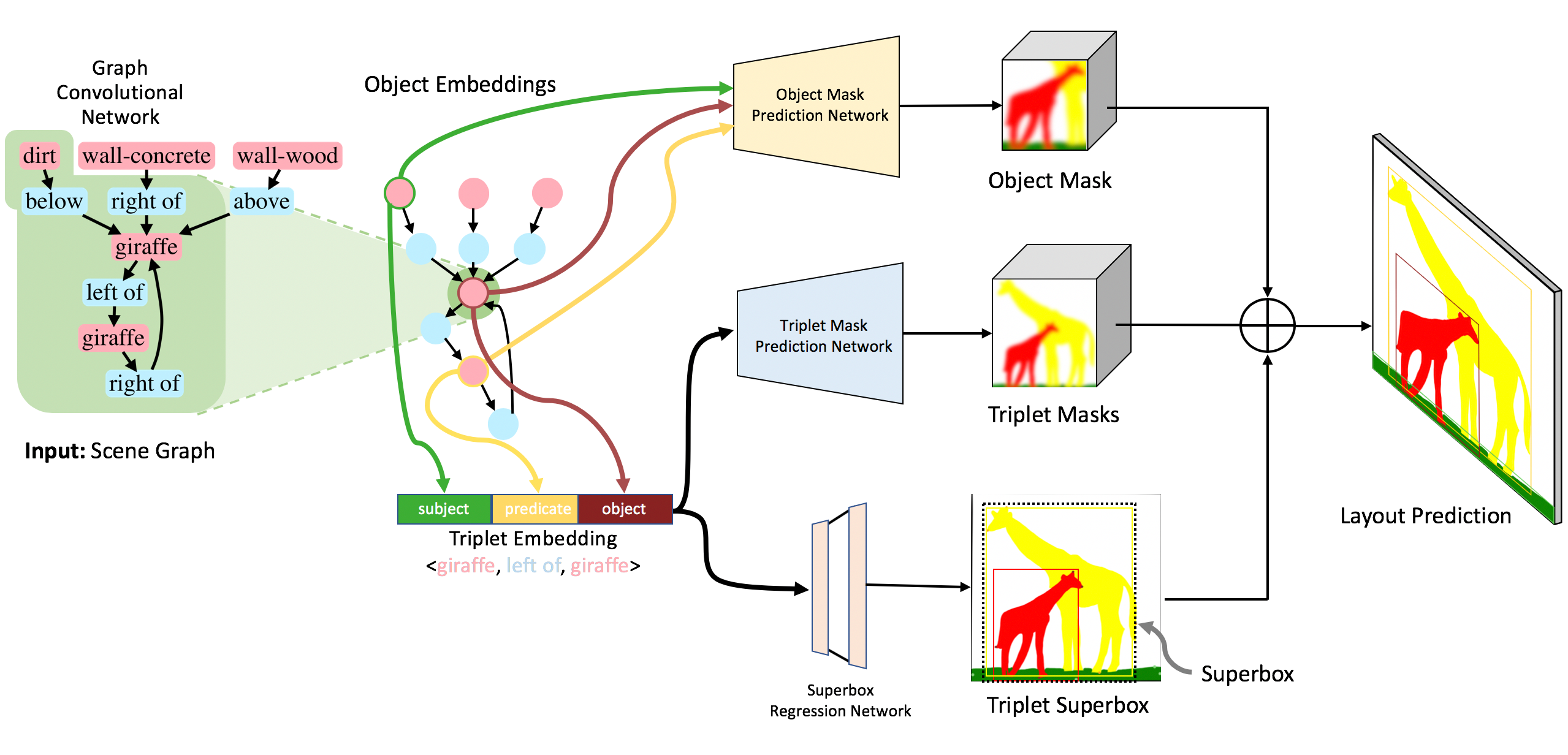}
\caption{\textbf{Triplet-Supervised Layout Prediction Network.} Here we show a multi-stage network which takes as input a scene graph and outputs a scene layout prediction. A graph convolutional neural network processes the scene graph to produce object embeddings, which can be used to form triplet embeddings composed of \textit{$<$subject, predicate, object$>$}. Both single object and triplet embeddings are passed to the next stage to predict object masks, triplet masks and triplet superboxes. Ultimately, these are combined to predict a scene layout.}
\label{fig:network}
\end{center}
\end{figure*}

\textbf{Layout Prediction with Triplet Supervision}. As part of our scene graph embedding introspection, we use a layout prediction network inspired by the image generation pipeline in \cite{img_gen_from_scene_graph18}. Figure \ref{fig:network} gives an overview of the network architecture. A GCNN processes an input scene graph to produce embeddings corresponding to object nodes in the graph. Singleton object embeddings are passed to the next stage of the layout prediction network per \cite{img_gen_from_scene_graph18}.

\begin{figure}[ht]
\begin{center}
\includegraphics[width=0.5\textwidth]{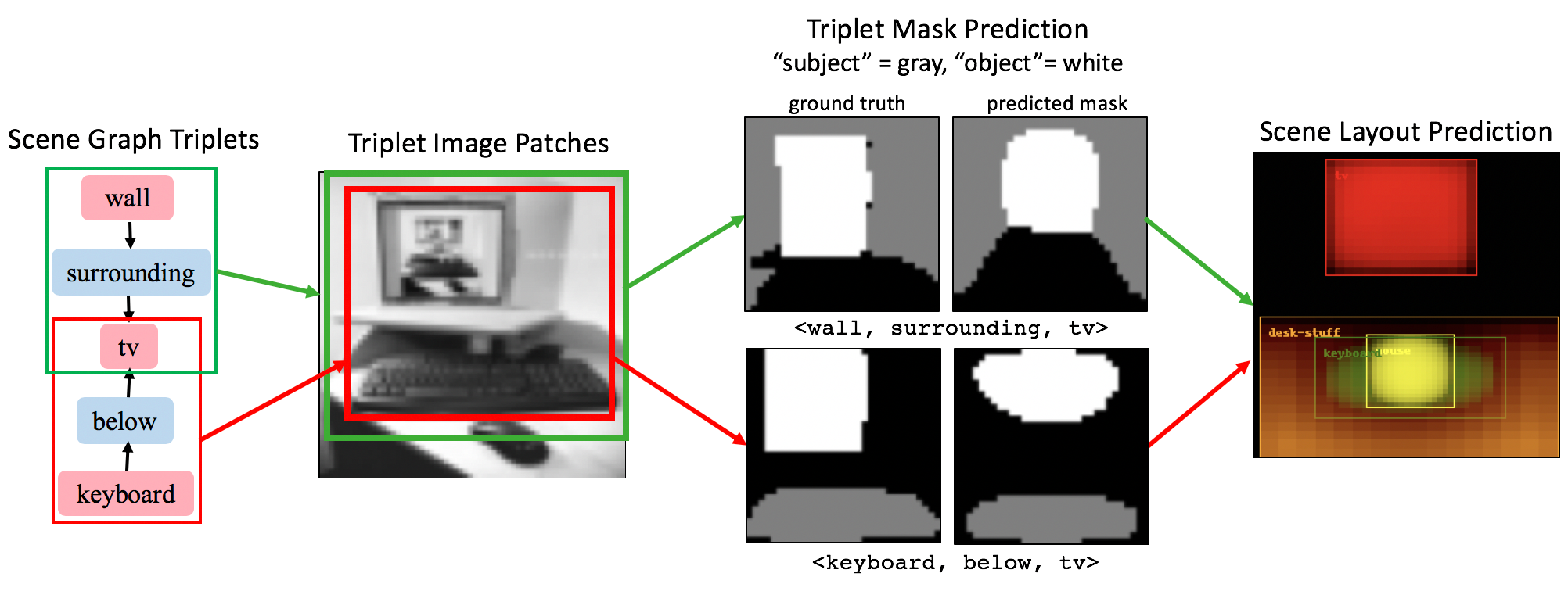}
\caption{\textbf{Triplet Mask Prediction.} Triplets containing a \textit{$<$subject,predicate,object$>$} found in a scene graph are used to predict corresponding triplet masks, labelling pixels either as subject and object. The mask prediction is used as supervisory signal during training.}
\label{fig:triplet_diagram}
\end{center}
\end{figure}

We utilize the object embeddings to form a set of triplet embeddings where each is composed of a \textit{$<$subject, predicate, object$>$} embedding.  We pass these through a triplet mask prediction network. Figure \ref{fig:triplet_diagram} highlights the details of the triplet mask prediction process.  Rather than just learn individual class labels, the network learns to label objects as either ’subject’ or ’object’, enforcing both an ordering and relationship between objects.  We also pass triplet embeddings through a triplet superbox regression network, where we train the network to do  joint localization over subject and object bounding boxes.

Ultimately, all of the outputs of the second stage of the layout prediction model are used to compose a scene layout mask with object localization.

\textbf{Data Augmentation}. We applied the method in \cite{ICLR_LLD} which uses heuristics to augment scene  graphs  with  new  spatial relations that induce a richer learned representation. The depth order between objects from an observer's viewpoint is quasi-exhaustively determined, and for 2D images, determining this order is non-trivial.   Linear perspective-based heuristics are utilized for augmenting spatial relationship vocabulary. 

\textbf{Training}. We train the layout prediction network to minimize two additional triplet-based losses in addition to those used in \cite{img_gen_from_scene_graph18}:
\begin{itemize}
  \item Triplet mask loss, $L_{triplet-mask}$, penalizing differences between ground truth triplet masks and predicted triplet masks with pixelwise cross-entropy loss.
  \item Triplet superbox loss, $L_{superbox}$, penalizing the $L_2$ difference between the ground truth and predicted triplet superboxes.
\end{itemize}

\section{Experiments}

\label{sec:results}

\subsection{Dataset}

\textbf{COCO}: We performed experiments on the 2017 COCO-Stuff dataset  \cite{cocostuff_caesar2018cvpr},  which  augments  
the  COCO dataset \cite{Lin2014MicrosoftCC} with additional stuff categories. The dataset annotates 40K train and 5K val images with bounding boxes and  segmentation  masks  for 80 thing categories  (people, cars, etc.)  and 91 stuff categories (sky, grass, etc.).  Similar to \cite{img_gen_from_scene_graph18}, we used thing and stuff annotations to construct synthetic scene graphs based on the 2D image coordinates of the objects, encoding six mutually exclusive geometric relationships:  \emph{left of, right of, above, below, inside, surrounding}. We ignored objects covering less than 2\% of the image, and used images with 3 to 8 objects.

\begin{table}[h]
\textbf{\caption{\textbf{Layout Prediction Model.}}}
\small
\begin{center}
\begin{tabular}{lcc} 
\textbf{Model} & \textbf{mIoU} & \textbf{Relation Score} \\
\hline
{\textbf{Baseline} \cite{img_gen_from_scene_graph18} } & 49.2\% & 54.1\% \\
{\textbf{Triplet Supervision}} & 50.3\% & 59.4\% \\
{\textbf{Triplet Supervision + DA}} & \textbf{52.3}\% & \textbf{61.7}\% \\ 
\hline
\end{tabular}
\end{center}
\vspace{1mm}
\label{table:layout_results}
\end{table}

\subsection{Scene Layout Prediction}

In order to evaluate the quality of scene layout prediction, we compared the predicted layout with the ground truth using intersection-over-union (mIoU) and relation score. The relation score measures the percentage of scene graph relations that are satisfied in the generated layout. The additional supervision and data augmentation introduced significantly improves over the baseline model \cite{img_gen_from_scene_graph18} as can be seen in Table 1. 


\subsection{Embedding Visualization and Separability} 

\begin{figure*}[ht]
\begin{center}
\includegraphics[width=1.0\textwidth]{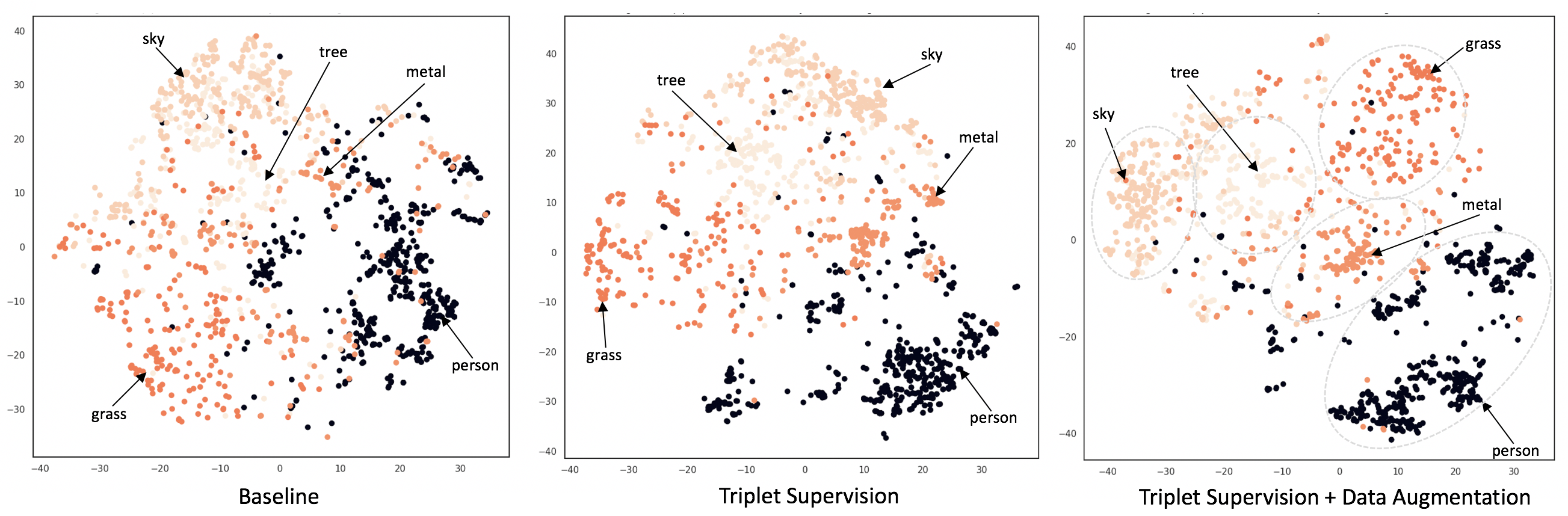}
\caption{\textbf{Triplet Supervision and Data Augmentation.} A t-SNE plot of the top 5 highest frequency classes in the COCO-Stuff dataset which shows  (from left to right) that as supervision and data augmentation are added, clusters representing each class are tighter and more distinct.}
\label{fig:top5_cluster}
\end{center}
\end{figure*}

To visualize the results, we first visualized the scene graph embedding using a t-SNE plot \cite{Maaten08visualizingdata}   (Figure \ref{fig:top5_cluster}) for the three models. For clarity, we only show the five most frequent classes in COCO-stuff dataset in the visualization. With the triplet supervision and data augmentation, the individual classes become more separable, as seen in the progression of models. We quantified this observation by using an SVM with a linear kernel to classify the class categories (Table 2). The classification accuracy of the ten most frequent classes along with the mean classification accuracy across all classes is shown. The model trained with triplet supervision and data augmentation shows significantly better class separability over the baseline, as measured by classification accuracy. This is especially apparent in lower frequency classes such as wall, building and pavement.  

\begin{figure}[ht] 
\begin{center}
\includegraphics[width=0.5\textwidth]{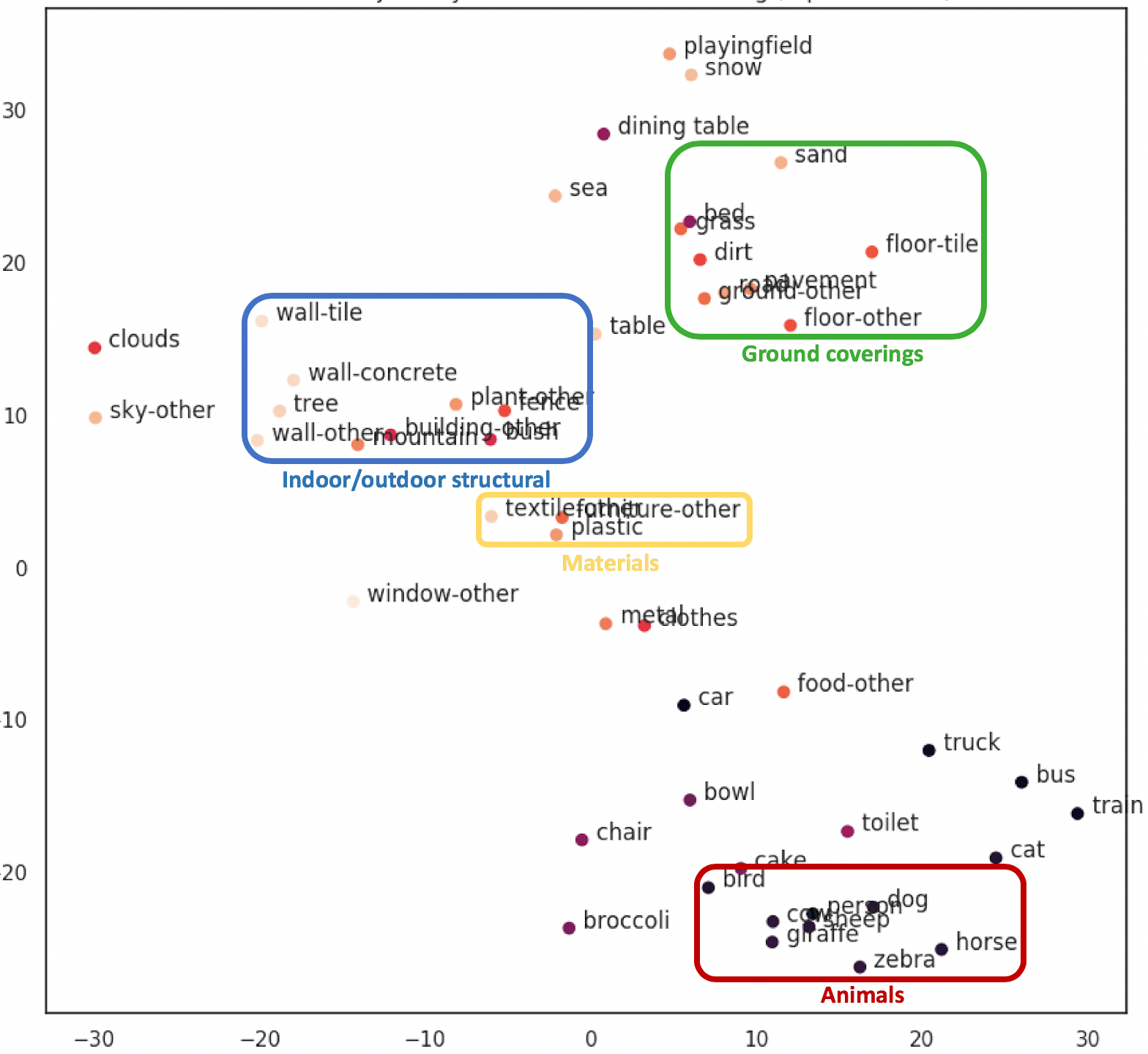}
\caption{\textbf{Mean Embedding Representation.} A t-SNE plot of mean embeddings for the top 50 highest frequency classes in the COCO-Stuff dataset. The embedding forms logical clusters of classes that are correlated.}
\label{fig:top50_mean}
\end{center}
\end{figure}

\begin{figure}[ht] 
\begin{center}
\includegraphics[width=0.5\textwidth]{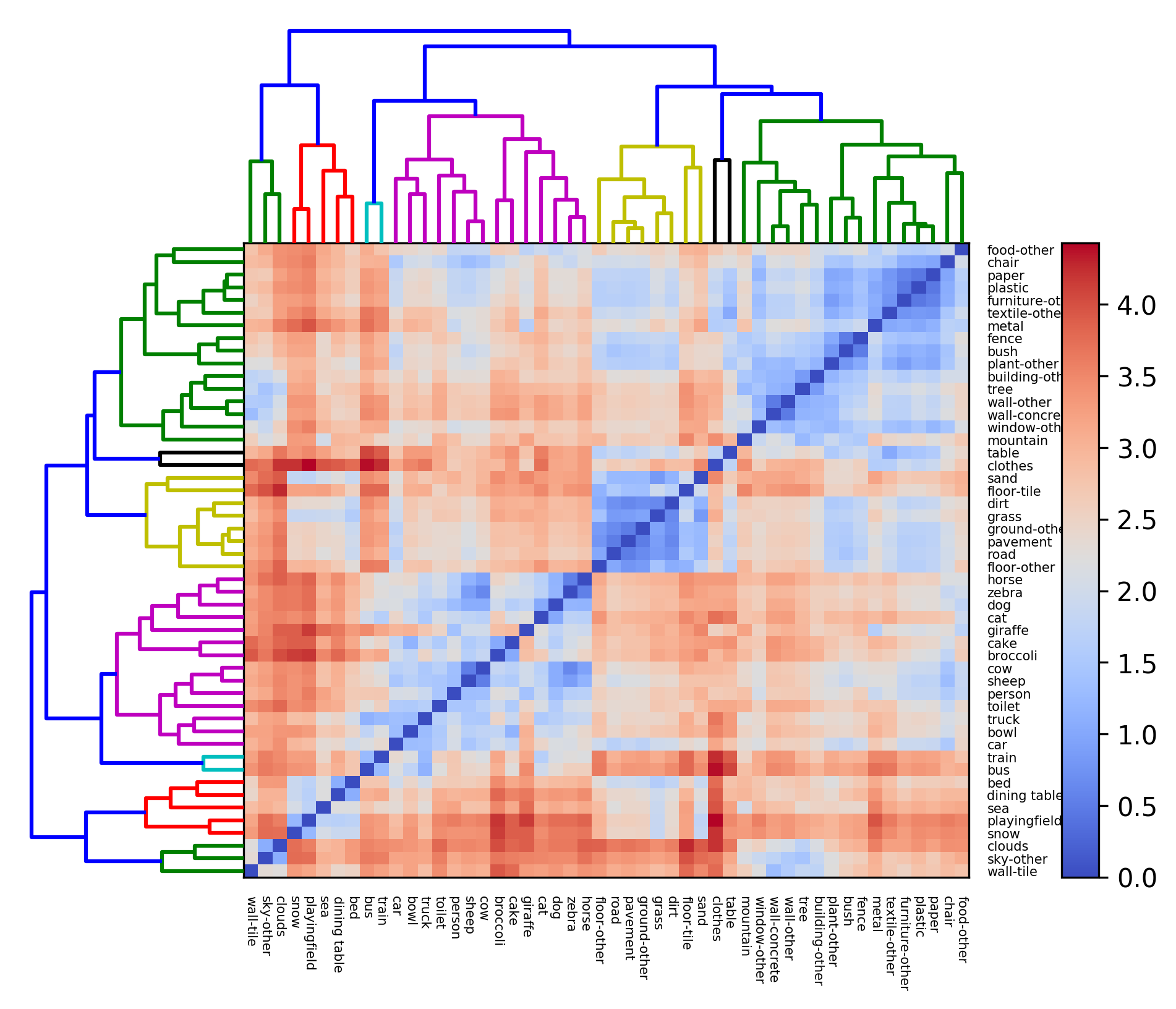}\caption{\textbf{Scene Graph Embedding Dendrogram} A dendrogram represented as distance heatmap between the mean embeddings of top 50 highest frequency in the COCO-Stuff dataset.\label{fig:top50_dendro}}
\end{center}
\end{figure}

In Figure \ref{fig:top50_mean}, the mean scene embedding nodes for the fifty most frequent classes in COCO-stuff are visualized. We can see that several semantically similar objects clusters form, such as ones containing ``animals" and ``ground coverings" (indicated by colored boxes).  Higher-level clusters such as ``indoor/outdoor structural" (blue box) and ``materials" (yellow box) which are  also semantically similar at a higher level, are found closer together in embedding space. This kind of hierarchical clustering can also be seen in the 2d dendrogram in Figure \ref{fig:top50_dendro} which visualizes a clustered heatmap of mean embedding distances between the top fifty most frequent classes. Distinct regions of blue (e.g. near-zero distance, other than along diagonal) indicate clusters of object classes that form correlated logical groupings in embedding space, such as outdoor objects (e.g. tree, bush, fence) and outdoor environments(e.g. sky, clouds, ground).

\begin{table*}[ht]
\caption{Scene Graph Embedding Classification}
\begin{tabular}{lccccccccccc}
\textbf{Model} & \textbf{person} & \textbf{tree} & \textbf{sky} & \textbf{grass} & \textbf{metal} & \textbf{wall} & \textbf{bldg} & \textbf{pavement} & \textbf{road} & \textbf{clothes} & \textbf{mean acc} \\ \hline
\textbf{Baseline} & 97.1\% & \textbf{75.6\%} & 85.9\% & 64.6\% & 77.6\% & 51.5\% & 30.6\% & 32.5\% & 52.3\% & \textbf{79.5\%} & 48.3\% \\
\textbf{Triplet Sup + DA} & \textbf{98.1\%} & 73.1\% & \textbf{89.4\%} & \textbf{70.8\%} & \textbf{84.8\%} & \textbf{61.9\%} & \textbf{50.9\%} & \textbf{59.1\%} & \textbf{58.8\%} & \textbf{79.5\%} & \textbf{59.4\%} \\ \hline
\vspace{1mm}
\end{tabular}
\label{table:layout_results}
\end{table*}

\section{Conclusion}
\label{sec:conclusion}

In the course of our scene graph embedding introspection, we have identified many key findings. We find that introducing triplet supervision and applying data augmentation significantly improves the performance of layout prediction models. We understand in turn that this correlates with improved scene embedding separability, quantified by measuring the classification accuracy of all nodes in the scene graph embedding. We have identified that scene graph embedding structure, sometimes expressed as separability, can be influenced by both method of training and richness of scene graph representation. We find overall there is a strong correlation between scene graph embedding structure and scene layout prediction model performance. The metrics for this kind of model (or other models trained with scene graph embeddings) may be useful in understanding future scene-graph related tasks.

{
\small
\bibliographystyle{ieee}
\bibliography{main}
}

\end{document}